\documentclass[10pt,twocolumn,letterpaper]{article}
\usepackage[accsupp]{axessibility}
\usepackage{iccv}
\usepackage{times}
\usepackage{epsfig}
\usepackage{graphicx}
\usepackage{amsmath}
\usepackage{amssymb}
\usepackage{algorithm}
\usepackage{algpseudocode}
\usepackage{pgfplots}
\pgfplotsset{width=8cm,compat=1.9}
\usepackage{subfig}
\usepackage{placeins}
\usepackage{colortbl}

\usepgfplotslibrary{external}
\usepackage[pagebackref=true,breaklinks=true,letterpaper=true,colorlinks,bookmarks=false]{hyperref}

\iccvfinalcopy 

\def\iccvPaperID{11080} 

\ificcvfinal\fi

\begin{document}

\title{CDFSL-V: Cross-Domain Few-Shot Learning for Videos}


\author{Sarinda Samarasinghe\\
{\tt\small sarinda.samarasinghe@ucf.edu}
\and
Mamshad Nayeem Rizve\\
{\tt\small mamshadnayeem.rizve@ucf.edu}
\and
Navid Kardan\\
{\tt\small nkardan@cs.ucf.edu}
\and
Mubarak Shah\\
{\tt\small shah@crcv.ucf.edu}\\
Center for Research in Computer Vision\\
University of Central Florida, Orlando, Florida, USA\\
}
\maketitle
\ificcvfinal\thispagestyle{empty}\fi

\begin{abstract}

Few-shot video action recognition is an effective approach to recognizing new categories with only a few labeled examples, thereby reducing the challenges associated with collecting and annotating large-scale video datasets. Existing methods in video action recognition rely on large labeled datasets from the same domain. However, this setup is not realistic as novel categories may come from different data domains that may have different spatial and temporal characteristics. This dissimilarity between the source and target domains can pose a significant challenge, rendering traditional few-shot action recognition techniques ineffective. To address this issue, in this work, we propose a novel cross-domain few-shot video action recognition method that leverages self-supervised learning and curriculum learning to balance the information from the source and target domains. To be particular, our method employs a masked autoencoder-based self-supervised training objective to learn from both source and target data in a self-supervised manner. Then a progressive curriculum balances learning the discriminative information from the source dataset with the generic information learned from the target domain. Initially, our curriculum utilizes supervised learning to learn class discriminative features from the source data. As the training progresses, we transition to learning target-domain-specific features. We propose  a progressive curriculum to encourage the emergence of rich features in the target domain based on class discriminative supervised features in
the source domain. 
We evaluate our method on several challenging benchmark datasets and demonstrate that our approach outperforms existing cross-domain few-shot learning techniques. Our code is available at \hyperlink{https://github.com/Sarinda251/CDFSL-V}{https://github.com/Sarinda251/CDFSL-V}

\end{abstract}

\section{Introduction}

%




\begin{figure*}[t]
\begin{center}
\includegraphics[width=1\linewidth]{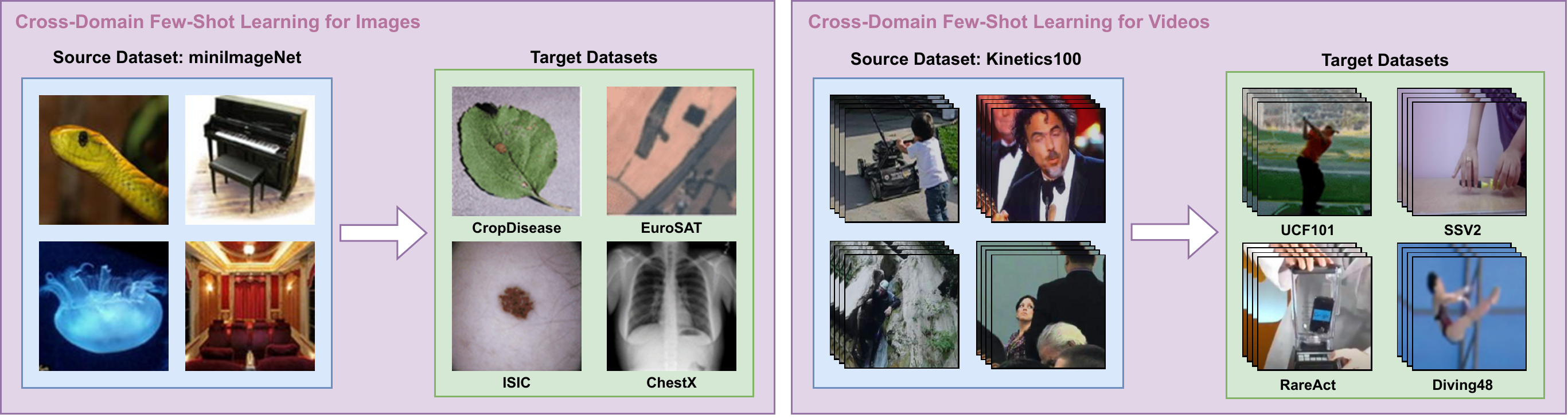}
\end{center}
\vspace{-2mm}
   \caption{On the left, we have the existing benchmark for CDFSL in the image domain. On the right, we present our proposed benchmark for CDFSL in the video domain. Our benchmark includes tasks from diverse target datasets, which require recognizing novel actions from different data distributions (UCF101, HMDB51), strong temporal reasoning (SSV2), atypical action understanding (RareAct), and fine-grained temporal understanding (Diving48).
   }
\label{fig:long}
\label{fig:onecol}
\end{figure*}

Even though deep learning is  inspired by the biological brain, in sharp contrast to humans, current deep models rely on large reservoirs of data to learn. The few-shot learning problem \cite{SurveyFSL} is introduced to close this gap, where a learning model should generalize solely based on a handful of training data. In traditional few-shot learning \cite{STARTUPtrans4}, the learning model is initially exposed to an annotated \emph{base dataset}, to learn generic features for the domain of interest. Then, this model is fine-tuned on a few labeled examples (support samples) of the test dataset and consequently evaluated on unlabeled test examples (query samples). However, this classic pipeline assumes the base and test datasets are from the same domain, thus closely related~\cite{STARTUPtrans1}.   

To mitigate this shortcoming, cross-domain few-shot learning (CDFSL) was proposed in \cite{BS-CDFSL}, where the base dataset is from a different domain than the test data. Interestingly, it is shown in \cite{UnderstandCDFSL} that standard transfer learning--- consisting of pre-training on the base dataset and fine-tuning on test data--- can significantly outperform few-shot learning methods in the cross-domain few-shot learning problem. Recently,  extra unlabeled test examples were incorporated in addition to the base dataset in \cite{STARTUP, DynamicDistillation} . Their approaches push forward cross-domain few-shot learning performance. In this paper, we follow this recent adaptation of CDFSL. 

While few-shot learning is widely studied in the computer vision community \cite{SummaryFSL}, video few-shot learning is less explored \cite{otam}. To the best of our knowledge, current methods in cross-domain few-shot learning are solely focused on image data. In this work, for the first time, we study cross-domain few-shot learning in the video domain. A common scheme in video few-shot learning~\cite{VideoFS}  utilizes an implicit assumption about video data, such as: a common mode of variation, similar temporal dynamics, or class distinctive features. However, in cross-domain few-shot learning, the base dataset can be drastically different from the target data, For instance, the RareAct dataset~\cite{RareAct} contains atypical actions which significantly deviate from the common actions present in the standard video datasets in terms of spatio-temporal dynamics, and the Diving48 dataset~\cite{Diving48} contains temporally fine-grained actions which have very similar spatial layout. Therefore, it is challenging to apply standard video few-shot learning methods to these datasets.









In the context of cross-domain few-shot learning, {\em supervised} pre-training on the source dataset has emerged as a common first step for most techniques \cite{STARTUP,DynamicDistillation}. This is because a strong source backbone can significantly contribute to the overall performance of the model \cite{UnderstandCDFSL}. However, simply relying on supervised pre-training may not be sufficient, especially when the target domain is substantially different from the source domain. To address this, in this work we propose to perform {\em self-supervised} pre-training on both source and target data to learn generic features. To be particular, we use recently proposed masked auto-encoder based \cite{VideoMAE} feature learning to learn generic features which are highly scalable and show better generalization performance. Nevertheless, the challenge remains on how to balance the learning of generic features (from source and target domain) and class discriminative features from the source dataset.

To this end, we propose a curriculum learning scheme by designing a progressive curriculum that balances learning the discriminative information from the source dataset with the generic information learned from the target domain. In the initial phase of the training, our curriculum utilizes supervised cross-entropy loss to learn class discriminative features from the source data. As the training progresses,  we strive to transition to the target domain through learning discriminative features in the target domain. 
To achieve this, we devise a schedule that increases the weight of a consistency loss to help with this transition. We  conduct extensive experiments to demonstrate the effectiveness of our proposed approach on various benchmark datasets. Our experiments  show significant improvements in cross-domain few-shot action recognition performance.

In summary, our work makes the following major contributions,

\begin{itemize}
    \item We propose a new, challenging, and realistic problem called cross-domain few-shot learning in videos (CDFSL-V). 
    \item We propose a novel solution based on self-supervised feature learning and curriculum learning for this challenging problem, which can address the difficulties associated with CDFSL-V by striking a balance between learning generic and class-discriminative features.
    \item We conduct extensive experimentation on multiple benchmark datasets. Our proposed method outperforms the existing methods in cross-domain few-shot learning, as well as, strong baselines based on transfer learning. 
\end{itemize}







\section{Related Work}

\textbf{Few-Shot Classification} Few-shot Learning methods can be split into two main categories: Meta-Learning and Transfer Learning \cite{LearnFromFew}. Meta-Learning \cite{LearnFromFewMeta7,LearnFromFewMeta8} framework provides a very common technique for few-shot learning algorithms, where the training procedure mimics the evaluation procedure. Just as few-shot evaluation consists of multiple few-shot episodes on the target test set, meta-learning techniques train a model in an episodic fashion on a meta-train set. In meta-learning this is done to encourage fast adaptation on the meta-test set. The other main approach in few-shot learning is Transfer Learning, where a model is pretrained on the source dataset before being fine-tuned on the target data for few-shot evaluation \cite{STARTUPtrans1,STARTUPtrans2,STARTUPtrans3,STARTUPtrans4}. Methods that use transfer learning aim to leverage as much information as possible from the source dataset in order to produce easily transferrable features to be adapted to the target dataset. Both methods assume some degree of similarity between the source and target datasets, hinging on the idea that features that can discriminate classes in the source domain can also discriminate classes in the target domain. When moving from images to videos, the introduction of temporal information adds to the difficulty of the task. OTAM \cite{otam} uses temporal alignment to improve few-shot classification for videos, using a distance metric to compare frames of the queries and the support set. STRM \cite{STRM} introduces a spatio-temporal enrichment module to look at visual and temporal context at the patch and frame level. HYSRM \cite{HYSRM} uses a hybrid relation model to learn relations within and across videos in a given few-shot episode. Our method focuses on training an encoder with generalizable features by leveraging unlabeled target data during training through both self-supervised learning and enforcing a consistency loss moderated by curriculum learning. 

\textbf{Self-Supervised Learning} Self-supervised learning has been shown to improve performance when combined with supervised learning by creating more transferable features \cite{VideoMAE}. These more generalizable features are extremely important in the cross-domain few-shot classification task, due to the domain gap and scarcity of labels. For self-supervised video classification, existing methods use contrastive learning to improve learning visual representation, at the cost of increased data augmentation and batch sizes \cite{videomae79,videomae45,videomae83,videomae5}.   Masked auto-encoders \cite{mae} mask patches of an image and attempt to reconstruct the missing parts. VideoMAE \cite{VideoMAE} extends this to video by adding space-time attention via a ViT backbone, providing a data efficient solution so the self-supervised video pretraining. We use VideoMAE as the backbone of our method.  

\textbf{Curriculum Learning} Curriculum learning involves prioritizing easier samples (or tasks) during training before increasing the weights of the more difficult samples \cite{Curriculum}. Typically, training examples are sorted by a difficulty metric, and used to create mini-batches of increasing difficulty for training the model \cite{PowerOfCurriculum}.  This method has shown success when applied in computer vision, specifically when used with transfer learning \cite{CurriculumTransfer}. 

For our problem setup we work with two datasets,the labeled source dataset and the unlabeled target data (for which we generate pseudo-labels), simultaneously during training. In our method we leverage curriculum learning such that we focus on the large labeled source dataset at the beginning of training, and eventually shifting towards equal weighting of the source and target losses.

\textbf{Cross-Domain Few-Shot Learning}
Similar to open-world semi-supervised learning \cite{cao2021open,guo2022robust,rizve2022towards,rizve2022openldn} that allows semi-supervised learning methods to perform on loosely related domains, the cross domain few shot learning framework permits base and test data that belong to different domains. BS-CDFSL \cite{BS-CDFSL} introduces a benchmark for the Cross-Domain Few-Shot problem for images. It consists of miniImageNet as the source dataset, and four target datasets of increasing difficulty: CropDisease \cite{CropDisease}, EuroSAT \cite{EuroSAT}, ISIC \cite{ISIC}, and ChestX \cite{ChestX}. STARTUP \cite{STARTUP} attempts to solve this problem by learning a teacher model on the source dataset that is  applied to generate pseudo-labels for the target dataset. Eventually, a new model on both the labeled source set and pseudolabeled target set is trained. Dynamic Distillation \cite{DynamicDistillation} improves upon this by updating the teacher model as a moving average of the student's weights. Both of these methods exhibit redundancy in the supervised training across their stages that we strive to eliminate in our approach.

While 
source-target dataset pairs such as UCF-HMDB51 from the SDAI Action II dataset \cite{SDAI} and the UCF-OlympicSport datasets \cite{UCFOlympic} have been proposed \cite{PASTN}, these dataset pairs share classes across domains, which  is not representative of the CDFSL problem. We take inspiration from the BS-CDFSL benchmark and use Kinetics-100 \cite{kin100} as our source, with UCF101 \cite{UCF101}, HMDB51 \cite{HMDB51}, Something-SomethingV2 \cite{SSv2}, Diving48 \cite{Diving48}, and RareAct \cite{RareAct} as our datasets. We ensure that we remove any class overlap between the source and target datasets.

\section{Methodology}
This section elaborates on our approach to tackle the CDFSL problem in the video domain. At the core of our method, we learn features from the source and target data in a supervised and self-supervised fashion, respectively. Furthermore, we propose a progressive curriculum to encourage the emergence of rich features in the target domain based on class discriminative supervised features in the source domain. In the following, first, we discuss our problem formulation (Sec.~\ref{sec:problem_formulation}). After that, we present our approach involving self-supervised feature learning and curriculum learning (Sec.~\ref{sec:approach}). 
\subsection{Problem Formulation}
\label{sec:problem_formulation}
The Cross-Domain Few-Shot Video Classification task requires the classification of an unlabeled query video belonging to the target dataset $\mathcal{D}_{T}$. A large labeled source dataset $\mathcal{D}_{S}$ is available during training. $\mathcal{D}_{S}$ and $\mathcal{D}_{T}$ have no shared classes, and usually have a significant domain gap. The unlabeled training split of $\mathcal{D}_{T}$ is leveraged during training, denoted as $\mathcal{D}_{T_U}$. For evaluation, multiple Few-Shot episodes are sampled from the testing split of $\mathcal{D}_{T}$. These episodes consist of a small labeled support set $\mathcal{S} \subset {D}_{T}$, consisting of a few labeled samples of each target class in the episode, and a disjoint query set $\mathcal{Q} \subset {D}_{T}$ to be classified. In the $N$-way $K$-shot classification setting, $\mathcal{Q}$ and $\mathcal S$ share the same $N$ classes sampled from $\mathcal{D}_{T}$ with $\mathcal S$ having $K$ labeled examples for each class.

\subsection{Approach}
\label{sec:approach}

\subsubsection{Self-Supervised Feature Learning} 

A fundamental challenge in solving few-shot problems is learning generalizable representations. A successful representation learning method is based on self-supervised learning, therefore, it has been readily applied to few-shot learning problem. Even though, it has yet to be applied in CDFS learning. Following the success of VideoMAE, to extract strong representations from video data we  apply VideoMAE model in our Pretraining phase. To this end, a rich set of features are extracted from a combination of the source and unlabeled-set of target dataset $\mathcal{D}_{S} \bigcup \mathcal{D}_{T_U}$. After this step, the encoder model from VideoMAE, $f$, is utilized as our primary feature extractor.

\subsubsection{Curriculum Learning}
Next, in our framework, we further improve the quality of the extracted features with the help of the ground-truth labels of the source data. To this end, we train a classifier $g$ on top of $f$, where this classifier outputs the number of classes equal to the classes in the source domain. Training a classifier in such a supervised manner makes the self-supervised representation more compact and class discriminative, particularly in the source domain. Ideally, we want to achieve the same in the target domain. However, doing such is difficult without accessing the ground-truth labels in the target domain. To overcome this challenge and to better utilize the target data, we minimize a consistency loss for the unlabeled target samples. This consistency loss is minimized at the output space of the source domain where pseudo-labels are generated using a teacher network.



\paragraph{Supervised Representation Learning}

To extract the discriminative features from the source dataset, we first train a student model $f_s$ based on a supervised loss on the labeled source data. We use the commonly used cross-entropy loss as the supervised loss, $\mathcal{L}_{sup}$, defined in the following,
\begin{align}
    \mathcal{L}_{sup} =& \mathcal{L}_{CE}(\mathrm{Softmax}(f_s(\mathbf{x}_i)), \mathbf{y}_i) \notag \\
    =& -\sum_{i=1}^M\mathbf{y}_i\log(\mathrm{Softmax}(f_s(\mathbf{x}_i))),
\end{align}
\noindent where, $\mathbf{x}_{i}\in\mathcal{D}_{S}$, $M=|\mathcal{D}_{S}|$, and $\mathbf{y}_i$ is the ground-truth label. The learned discriminative features provides us with more generalizable features to the target domain.  

\paragraph{Unsupervised Representation Learning}
For the unlabeled data from target domain, we apply pseudo-labels to increase generalizability of the learned features in an unsupervised fashion.
To this end, after obtaining the pseudo-labels we compute a consistency loss. The consistency loss ensures that the representations from the student model match the representations from a teacher network. We create a teacher model $f_t$ by taking an exponential moving average of the student model in the following manner,
\begin{align}
    f_t^{(i+1)} = \alpha f_t^{(i)} + (1-\alpha)f_s^{(i+1)},
\end{align}
\noindent where, $\alpha$ is the exponential decay parameter, $i$ refers to $i$th iteration. 


This consistency loss ensures that the $f_s$ predictions for unlabeled target data match with the pseudo-labels generated from $f_t$. 
Additionally, following the success of DINO \cite{DINO} we want to extract features that can learn a local-to-global  relationship between data. To this end, each batch of unlabeled target data $\mathbf{X} \in \mathcal{D}_{T_U}$, is transformed into two separate sets to make strong and weak augmented copies of the batch: $\mathbf{X}_{str}$ and $\mathbf{X}_{weak}$. To be specific, we use temporally consistent  $\mathrm{RandomResizeCrop}$ and $\mathrm{RandomHorizontalFlip}$ as a set of weak augmentations, while the set of strong augmentations consists of temporally consistent $\mathrm{RandomColorJitter}$, $\mathrm{RandomGreyscale}$, and $\mathrm{RandomGaussianBlur}$ in addition to the set of weak augmentations. 

To compute the consistency loss, first the weakly augmented unlabeled target data is passed through the teacher model to get the teacher outputs $f_t(\mathbf{X}_{weak})$. These outputs are then sharpened by a temperature $\tau$ to form pseudo-labels for the target data after performing the $\mathrm{Softmax}$ operation.  The consistency loss is a cross-entropy loss between the student outputs of the strongly augmented videos $f_s(\mathbf{X}_{str})$ and the sharpened teacher outputs which is defined in the following, 
\begin{align}
    \mathcal{L}_{con} = -\sum\mathbf{\hat{Y}}\log(\mathrm{Softmax}(f_s(\mathbf{X}_{str}))),
\end{align}
\noindent where, $\mathbf{\hat{Y}}=\mathrm{Softmax}(f_t(\mathbf{X}_{weak})/\tau)$.

The overall training objective for updating the parameters of the student network is a weighted average of the Supervised and the Consistency losses, defined in the following,
\begin{align}
    \mathcal{L}_{total} = \mathcal{L}_{sup} + \lambda\mathcal{L}_{con},
\end{align}
\noindent where, the consistency loss scaling parameter $\lambda$ controls the relative contribution of consistency loss to the total loss.

While previous CDFS methods have applied both supervised loss and consistency loss, they applied them in separate stages~\cite{DynamicDistillation,STARTUP}. One of the unique characteristics of our approach is to combine these losses through curriculum learning, which not only simplifies the training pipeline but also improves performance.

In our curriculum, we adjust the difficulty of the consistency through tuning its scaling parameter $\lambda$ following a pre-defined curriculum. In particular, at the beginning of training, we set the consistency loss scaling parameter, $\lambda$, to a very low value. This makes the beginning of the training similar to performing supervised training solely on the source dataset. As the training progress, we emphasize the importance of consistency by increasing $\lambda$ over the course of the training, which encourages the emergence of local-to-global features that can potentially generalize better in the target domain.  
Additionally, to facilitate the transition from the source domain to the target domain, we decay the learning rate of the classifier in the student model over the course of training. Initially, this classifier is trained at the same rate as the rest of the student model. This learning rate is decreased over the course of training to emulate freezing the classifier after supervised training on the source data.
Once the training is complete, the student model is kept and the classifier is discarded. Using the labeled support set of the target data, a new logistic regression layer $c'$ is learned on top of the student model. The model can now be used for inference on the target query images. The entire procedure is summarized in Algorithm \ref{alg:cap}.

\begin{figure*}[t]
\begin{center}

\includegraphics[width=0.8\linewidth]{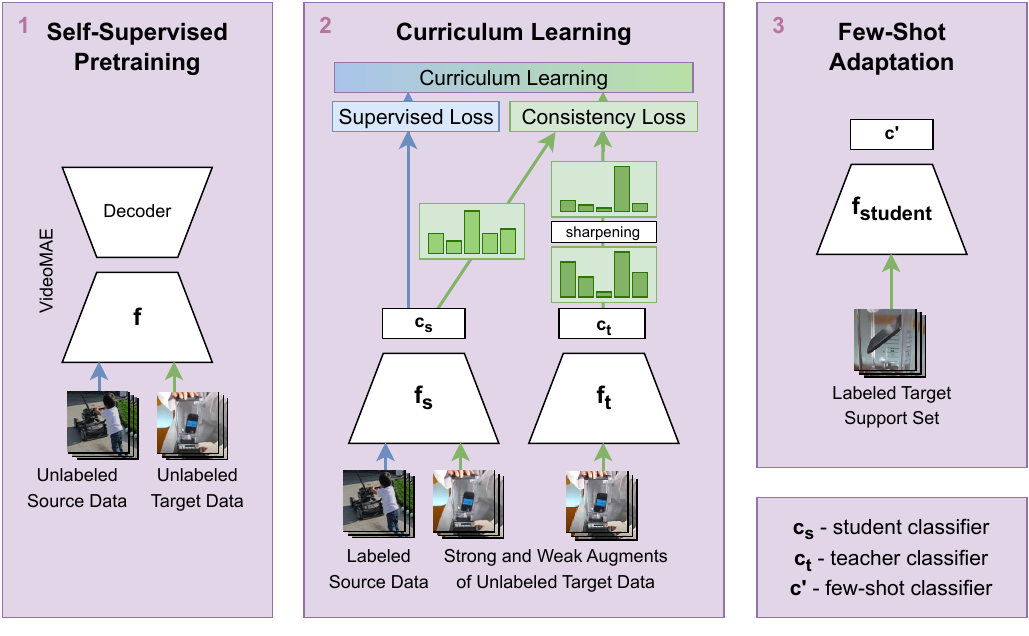}

\end{center}
\vspace{-4mm}
   \caption{Our goal is to solve the cross-domain few-shot learning task for the target dataset, leveraging the labeled base dataset alongside unlabeled target data. Our method three three stages: {\bf 1}:  Self-supervised pretraining of an autoencoder on both the source and target data without labels is performed. {\bf 2}: The encoder is used to initialize a student and teacher model for curriculum learning. We compute a supervised loss on the labeled source data. For the consistency loss, we generate pseudo-labels using the sharpened teacher output for weakly augmented target images. The pseudo-labels are then used with the student output on strong augmentations of the same images to calculate the consistency loss. The supervised and consistency losses are both used to directly update the student, while the teacher is updated as a moving average of the student's weights. {\bf 3}: for few-shot evaluation, the student classifier is replaced with a few-shot classifier that is fine-tuned on the labeled target support set. This classifier can then be used to classify the target query images.}
\label{fig:long}
\label{fig:onecol}
\end{figure*}
\begin{algorithm}
\caption{Curriculum Learning for CDFSL-V}\label{alg:cap}
\begin{algorithmic}
\State $f_s$, $f_t$: student, teacher model with parameters $\theta_s$, $\theta_t$.
\State $\tau$: teacher temperature
\State $\alpha$: momentum rate to update teacher
\For {($\mathbf{x}_s$, $\mathbf{y}_s$), $\mathbf{x}_t$ in loader}
\State sample $\mathbf{x}_s, \mathbf{y}_s$ from base data
\State sample $\mathbf{x}_t$ from target data
\State $\mathcal{L}_{sup} = \mathcal{L}_{CE}(f_{s}(\mathbf{x}_s), \mathbf{y}_s)$
\State $\mathbf{x}_{weak}, \mathbf{x}_{str} = \mathrm{WeakAug}(\mathbf{x}_t), \mathrm{StrongAug}(\mathbf{x}_t)$
\State $out_t$, $p_s = f_{t}(\mathbf{x}_{weak}), \mathrm{Softmax}(f_{s}(\mathbf{x}_{str}))$
\Comment{teacher logits and student pseudo-labels}
\State $p_t = \mathrm{Softmax}(out_t / \tau, dim=-1).detach()$
\Comment{sharpen + stop-grad}
\State $\mathcal{L}_{con}=\mathcal{L}_{CE}(p_s, p_t)$ \Comment{consistency loss}
\State $\mathcal{L}_{total} = \mathcal{L}_{sup} + \lambda\mathcal{L}_{con}$
\State $\theta_s\leftarrow\theta_s+\beta\nabla_{\theta_s} \mathcal{L}_{total}$ \Comment{update student}
\State $\theta_t\leftarrow \alpha\theta_t+(1-\alpha)\theta_s$ \Comment{update teacher}
\EndFor
\end{algorithmic}
\end{algorithm}

\section{Experiments} 
 In this section we evaluate our proposed method with strong transfer learning baselines and the recent techniques applied in cross-domain few-shot learning. For a thorough comparison, we utilize a variety of target domains to capture performance of different methods when encountering a variety of cross-domain scenarios. Our main result is that our approach outperforms existing sate-of-the-art cross-domain few-shot learning techniques. Finally, we conduct an ablation and analyse the significance of different components of our approach.     

\subsection{Datasets}

We use the Kinetics-100 \cite{kin100} train split as our Source dataset. It contains 100 of the original dataset further split into 64, 12, and 24 class subsets for train, validation, and test, respectively for few-shot action recognition. We also conduct experiments on the larger Kinetics-400 \cite{kin400} (Table \ref{k400}). Due to class overlap between Kinetics and two of our target datasets, UCF101 and HMDB51, we remove the overlapping classes from the source dataset. Without this removal, the supervised training on shared classes between the source and target datasets would be an unfair representation of the Cross-Domain Few-Shot problem setting.
The target datasets in the order of increasing difficulty are: UCF101, RareAct, HMDB51, Something-SomethingV2, and Diving48. UCF101 and HMDB51 are most similar to Kinetics datasets in terms of domain gap. They even have overlapping classes that needed to be removed in order to make them appropriate target datasets However, that is not the case for the other target datasets. For instance, the Something-SomethingV2 dataset has 87 classes, consisting of actions doing `something' to `something'. This dataset primarily contains zoomed-in videos focusing on the object instead of the person which is generally not the case for actions present in the actor-centric Kinetics dataset. Diving48 on the other hand is a dataset for fine-grained action recognition with 48 different dives, each comprised of different sequences of complex sub-actions. The RareAct is very different from all other source and target datasets since it contains unusual actions like `blend phone' and is generally used for evaluating few/zero-shot action compositionality. For evaluation, we compute the 5-way 5-shot accuracy on the test-split for each target dataset.  


\subsection{Experiment Details}
We use the encoder network from VideoMAE with a ViT-S backbone for our feature extraction. For videos we sample 16 frames at a $112 \times 112$ resolution.
We train on the combined training data of both the source and target datasets without labels for $400$ epochs at a batch size of $32$ using SGD optimizer at a learning rate of $0.1$. After initializing the student and teacher models using the VideoMAE encoder, the student is trained for $200$ epochs on the combined supervised and consistency losses. The student is updated directly using SGD with a learning rate of $0.01$, and the teacher is updated as a moving average of the student weights with a momentum of $0.9$. The teacher output is sharpened at a temperature of $0.1$ to be used as pseudo-labels for the student output on the unlabeled target data. The batch size used for the curriculum learning stage is $16$. 
Over the course of training, we set the consistency loss scaling parameter, $\lambda$ as:
\setlength{\abovedisplayskip}{8pt}
\setlength{\belowdisplayskip}{8pt}
\setlength{\abovedisplayshortskip}{0pt}
\setlength{\belowdisplayshortskip}{0pt}
\begin{align}
    \label{consScaleEq}
    \lambda_{cons} = \frac{\arctan(10(x-.5))}{\pi} + .5
\end{align}
where $x$ is the ratio of current epoch to total epochs in training. This reduces the weight of the consistency loss significantly in the start of training, while making it on par with the supervised loss towards the end. Similarly, the learning rate for the student classifier head (the classifier layer following $f_s$) decayed according to $\lambda_{cls} = \frac{\arctan(-10(x-.5))}{\pi} + .5$.
so that the classifier head learns primarily from the supervised loss early on and effectively freezes towards the end of training. 

For few-shot adaptation, the student encoder is retained and the student classifier head is discarded. We then learn a new logistic regression classifier on top of the encoder using a sampled 5-way 5-shot support set from the target testing data. We report the accuracy on the remaining testing data for the selected classes.

\begin{figure}
    \setlength{\columnsep}{10pt}%
    \includegraphics[width=0.9\linewidth]{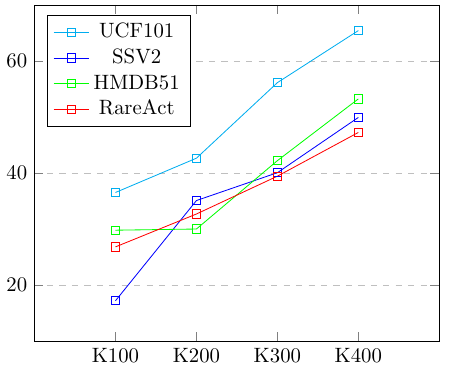}
    \caption{Results with varying size of source data.}
    \label{fig:var_source}
\end{figure}


\begin{table*}
\begin{center}
\resizebox{0.8\textwidth}{!}{
\begin{tabular}{|l|c|c|c|c|c|c|}
\hline
Method &  UCF101 & SSV2 & HMDB51 & Diving48 & RareAct & Average\\
\hline\hline
Random Initialization &23.83&16.02&12.08&15.37&16.57&16.78 \\
STARTUP++ &60.82&39.60&44.71&14.92&45.22&41.05 \\
Dynamic Distillation++ &63.26&44.50&48.04&16.23&47.01&43.81 \\
STRM &42.33&35.01&24.98&16.69&39.01&31.60\\
HYRSM &45.65& 40.09& 29.81& 17.57&44.27 & 35.49\\
\rowcolor[gray]{.95} \textbf{Ours} & \textbf{65.42}&\textbf{49.92}&\textbf{53.23}&\textbf{17.84}&\textbf{49.80}&\textbf{47.24} \\
\hline
\end{tabular}}
\end{center}
\vspace{-6mm}
\caption{5-way 5-shot Accuracy using Kinetics-400 as the source dataset. We use STARTUP++ and Dynamic Distillation++ to denote that these methods include self-supervised pretraining, despite being used in their original papers.}
\label{k400}
\end{table*}

\begin{table*}
\begin{center}
\resizebox{0.9\textwidth}{!}{
\begin{tabular}{|l|c|c|c|c|c|c|}
\hline
Method, Source Dataset: Kinetics-100&  UCF101 & HMDB51 & SSV2 & Diving48 & RareAct & Average\\
\hline\hline

Equal Loss Weighting &32.02&27.39&15.34&16.07&33.67&24.90 \\
No Temperature Sharpening & 34.01 & 28.18	&15.21&	16.77 &	33.80 & 25.59\\
Self-Supervised Training & 37.54 & 25.09 & 16.21 & 17.14 & 29.58 &25.11\\
Supervised Training & 32.06	& 23.86	& 14.40	& 16.16 & 31.15 &23.53\\
{\bf Ours} & {\bf 36.53}	& {\bf 29.80} & {\bf 17.21} & {\bf 16.37} & {\bf 33.91} & {\bf 26.82} \\
\hline
\end{tabular}}
\end{center}
\vspace{-4mm}
\caption{The effect of removing different components of our proposed method.}
\label{tab:ablation}
\vspace{-6mm}
\end{table*}

\subsection{Kinetics-400 Experiment}

Random initialization is used as the baseline for this experiment and entails learning a logistic regression classifier on top of an \emph{untrained} VideoMAE encoder. We compare our method to two Cross-Domain Few-Shot methods for images, as no other methods exist to solve the CDFSL problem for videos. For this experiment, we include self-supervised pre-training for Dynamic Distillation and STARTUP, denoting them as Dynamic Distillation++ and STARTUP++, respectively. In addition, we compare our method to two Few-Shot methods for Videos: STRM \cite{STRM} and HYRSM\cite{HYSRM}. Our method outperforms the previous state-of-the-art method, Dynamic Distillation~\cite{DynamicDistillation}, across all 5 target datasets while using the Kinetics-400 dataset as the source. Additionally, the absolute improvement in classification performance is consistent with the aforementioned relative difficulty of each of the target datasets, with Diving48 improving the least. 

Our main result is that we do better than existing CDFSL methods for images, as well as the Few-Shot methods for videos. As shown in Table~\cite{kin400}, We outperform Dynamic Distillation by 2.2\% on UCF101, 5.2\% on HMDB51, 5.4\% on SSV2, 2.7\% on RareAct, and 2.8\% on Diving48, averaging to a 3.4\% increase. Compared to STARTUP, STRM, and HYSRM our method outperforms by 6.19\%, 15.64\%, and 11.8\%, respectively. Interestingly, even our modified image baselines (STARTUP++ and Dynamic Distillation++) outperform these video few-shot methods, highlighting the inadequacy of traditional video few-shot approaches for this challenging Cross-Domain Few-Shot problem.


\subsection{Kinetics-100/200/300 Experiments}
We repeat the experiments using Kinetics-100, Kinetics-200, and Kinetics-300 as the source datasets. We compare our method's performance across these varying source datasets. In this experiment, we evaluate how the increase in the number of classes in the source dataset impacts performance. As shown in Fig~\ref{fig:var_source}, increasing the size of the source dataset consistently improves performance on all datasets.

\begin{figure*}
\begin{center}
\subfloat[HMDB51]{\includegraphics[width = 2in]{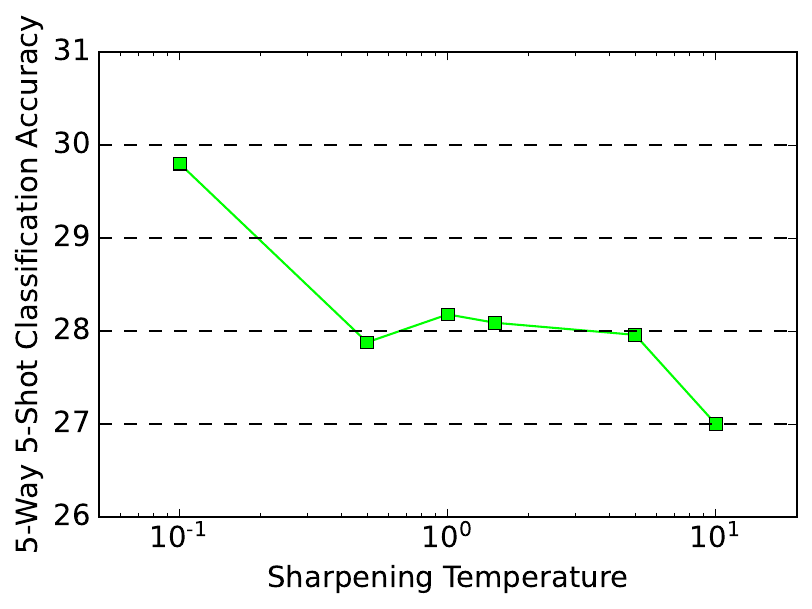}} 
\subfloat[Diving48]{\includegraphics[width = 2in]{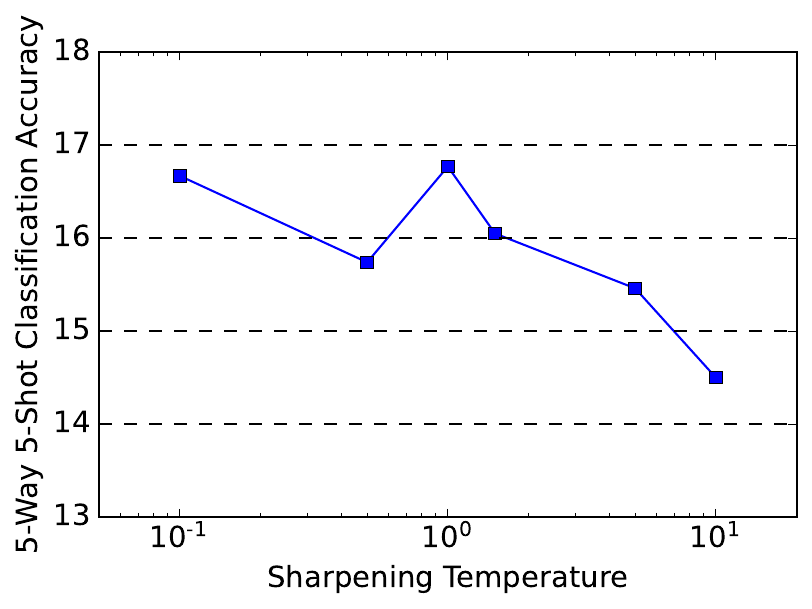}}
\subfloat[RareAct]{\includegraphics[width = 2in]{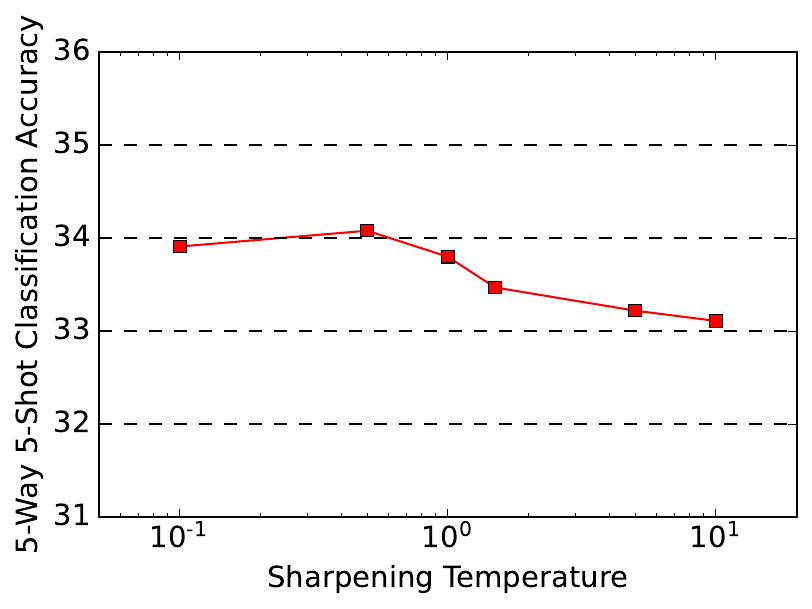}}
\vspace{-2mm}
\caption{Temperature parameter experiments. We use Kinetics-100 as the source dataset and vary the sharpening temperature for the teacher pseudo-labels. As the temperature increases (and sharpness decreases) the performance tends to decrease.}
\label{temp-sharp}
\end{center}
\vspace{-6mm}
\end{figure*}

\subsection{Ablation and Analysis}

In this section we analyze the importance of different components of our approach. Particularly we study the effect of increasing the size of the source dataset, the effect of sharpening temperature, and the impact of curriculum learning on the performance of the method.

\textbf{Increasing the size of the source dataset}
In few-shot learning and transfer learning literature, it is common to utilize a source domain with a significantly larger number of classes than the target domain \cite{BS-CDFSL}. A dataset with a larger number of classes can capture a more diverse set of features which facilitates its application on less diverse datasets. For example, in the BS-CDFSL benchmark for images, miniImageNet, the source dataset, has 100 classes. The target image datasets: CropDisease, EuroSAT, ISIC, and ChestX have $38$, $10$, $5$, and $15$ classes, respectively. In that setup, the source dataset has over double the amount of classes of the largest target dataset. In comparison, the source dataset in our video benchmark has $61$ classes, which is less than two of the target datasets: UCF101 with $101$ and Something-SomethingV2 with $87$. In this experiment, we explore the impact of the size of the source dataset in CDFSL. To be more specific, we apply the larger Kinetics-400 dataset instead of Kinetics-100 as the source.

Both STARTUP and Dynamic Distillation make use of supervised pretraining on the source dataset. For the experiments on Kinetics-400, we supplement the pretraining stages of both methods with self-supervised pretraining as well to highlight the effect of our curriculum-based schedule. In Table \ref{k400},
We observe a drastic improvement in the performance when we utilize a more diverse source dataset. Interestingly, we further notice an increase in the relative performance of our method to Dynamic Distillation.

\textbf{Temperature Sharpening Analysis}
 As in STARTUP \cite{STARTUP}, we want to leverage the unlabeled target data during training by using a consistency loss. We use the teacher model to create the ground truth for this loss, dividing the teacher output by the temperature parameter T to sharpen it and use as pseudo-labels. Similarly to Dynamic Distillation \cite{DynamicDistillation}, sharpening of the labels is used to develop low-entropy predictions from the student.
 
 We study the impact of temperature sharpening by setting the temperature parameter to $1$ (with the default value taken from Dynamic Distillation being $0.1$), making the teacher output completely unsharpened. As shown in second row of Table~\ref{tab:ablation}, removing the temperature sharpening reduces performance in almost all datasets (the exception being Diving48 with a $0.4\%$ increase) with an average decrease in performance of $1\%$ compared to our original method. We can see that temperature sharpening has a slight but positive impact when used with our CDFSL problem setup for videos.

 \textbf{Pretraining Baselines} 
 It has been shown that pretraining contributes a significant part to few-shot learning \cite{UnderstandCDFSL}. To examine how much of the performance is attributed to this, we compare some established transfer learning baselines with multiple pretraining configurations followed by few-shot adaptation. Self-supervised 
 training refers to only self-supervised training on the combined source and target datasets without labels, and supervised 
 training is simply training on the labeled source dataset. In rows 3 and 4 of Table~\ref{tab:ablation}, we can see the contribution of each of the pretraining techniques. For most of the datasets, only using self-supervised pretraining outperforms  using supervised pretraining, with the exception being RareAct. On average, our method performs 1.7\% over the self-supervised baseline and 3.3\% over the supervised baseline.

\textbf{Impact of Curriculum Learning}
The motivation behind curriculum learning is to ease the training of the model by focusing more on easier data first. For our problem setup where we leverage unlabeled target data alongside the labeled source, we begin with focusing more on the supervised source loss as it is an easier task than matching target videos to pseudo-labels in the source domain. Once the model has sufficiently learned relationships from the source dataset, the importance of the target consistency loss can increase to help improve the adaptation.

We use $\lambda_{cons}$ to scale the consistency loss during training, as shown in Eq.~\ref{consScaleEq}. To analyze the effect of enforcing the curriculum scaling, we compare  keeping $\lambda_{cons}$ at $1$ for the entirety of training and making both supervised and consistency losses weighted equally the whole time. Additionally, we train our model at temperatures of $0.5$, $1.5$, $5$, and $10$ as shown in Figure \ref{temp-sharp}. Weighting both losses equally results in an average drop in performance of $1.6\%$. We see that using curriculum learning improves the 
performance. 

\section{Conclusion}
In this paper, we addressed the problem of cross-domain few-shot action recognition in videos, which is a challenging and realistic problem with several practical applications in fields such as robotics. We  proposed a novel approach based on self-supervised feature learning and curriculum learning to address the challenges associated with this problem. Our approach strikes a balance between learning generic and class-discriminative features, which significantly improves the few-shot action recognition performance. We  conducted extensive experiments on various benchmark datasets, where our proposed method outperforms current cross-domain few-shot learning methods in the image domain and few-shot learning methods in the video domain. Our work contributes to the computer vision community by introducing a new problem and providing a novel solution to address it. We hope that this work will inspire further research in this direction and help advance the state-of-the-art in few-shot action recognition.

\section{Acknowledgements}
This research is based upon work supported in part by the Office of the Director of National Intelligence (IARPA) via 2022-21102100001. The views and conclusions contained herein are those of the authors and should not be interpreted as necessarily representing the official policies, either expressed or implied, of ODNI, IARPA, or the US Government. The US Government is authorized to reproduce and distribute reprints for governmental purposes notwithstanding any copyright annotation therein.

\FloatBarrier

{\small
\bibliographystyle{ieee_fullname}
\bibliography{egbib}
}

\end{document}